# A Case Based Reasoning Approach for Answer Reranking in Question Answering

Karl-Heinz Weis

Artificial Intelligence Working Group
University of Koblenz-Landau
Universitätsstr. 1
56016 Koblenz, Germany
khweis@weis-consulting.de

**Abstract:** In this document I present an approach to answer validation and reranking for question answering (QA) systems. A cased-based reasoning (CBR) system judges answer candidates for questions from annotated answer candidates for earlier questions. The promise of this approach is that user feedback will result in improved answers of the QA system, due to the growing case base. In the paper, I present the adequate structuring of the case base and the appropriate selection of relevant similarity measures, in order to solve the answer validation problem. The structural case base is built from annotated MultiNet graphs, which provide representations for natural language expressions, and corresponding graph similarity measures. I cover a priori relations to experienced answer candidates for former questions. I compare the CBR System results to current approaches in an experiment integrating CBR into an existing framework for answer validation and reranking. This integration is achieved by adding CBR-related features to the input of a learned ranking model that determines the final answer ranking. In the experiments based on QA@CLEF questions, the best learned models make heavy use of CBR features. Observing the results with a continually growing case base, I present a positive effect of the size of the case base on the accuracy of the CBR subsystem.

## 1 Introduction

Question answering (QA) systems aim at providing answers to questions.[1] The QA process is often divided into generation of answer candidates, using information retrieval techniques, followed by a validation and selection step that determines the final ranking. The ranking accuracy is crucial since users typically inspect only a few top-ranked results. In this paper I propose to extend the existing LogAnswer solution by a case-based reasoning (CBR) approach, where annotated answer candidates for known questions provide evidence for validating answer candidates for new questions. This promises a continuous increase in answer quality, if the QA system of interest is available online and users constantly provide feedback, thus extending the case base. I further propose the integration of graph similarity measures, so that the CBR approach will complement logical validation in comparing semantic structure of questions and potential answers. The idea of using CBR technology for question answering is not new. However, its application has typically been limited to FAQ answering, where questions are handled by spotting the intended question and its corresponding answer in a given list of question/answer pairs. Examples of such systems are [JP08], [BH97], [FC94], [LH98]. They rest on techniques for textual CBR [LA98]. By contrast, I aim at utilizing user annotations for earlier user questions for improving system responses of a regular open-domain QA system whose range of supported questions is in principle unlimited and cannot be restricted to a small fixed list. While FAQ answering systems use CBR technology for directly retrieving answers to the considered question, I propose also the use of CBR as a new tool for answer validation. The paper is organized as follows. Sect. 2 sketches the architecture and functioning of the LogAnswer QA system, which provides the test bed for the CBR extension. Sect. 3 introduces the semantic network formalism which provides the graph representations on which the CBR module must operate. Sect. 4 introduces the CBR approach for answer validation. It covers the adequate structuring of the case base and the appropriate selection of relevant similarity measures. Sect. 6 explains the experiments and their resuls..

---

[1] The annual CLEF systems evaluation campaign (http://www.clef-initiative.eu/) and the TREC initiative (http://trec.nist.gov/) feature (or have featured) QA tracks, and a wealth of papers on QA systems technology and evaluation are available from these websites.

## 2 System Overview

LogAnswer [FG10], [GP09] is an open-domain QA system for German that can be accessed on the web.[2] It works with a knowledge base derived from textual sources, namely the German Wikipedia and a corpus of newspaper articles. Answers are produced using a combination of deep linguistic processing and automated theorem proving [FG10]. About 12 million sentences have been parsed into semantic networks and stored using scheme syntax [FG10], using the WOCADI parser [Ha03] for linguistic analysis. Given a question, the system first retrieves pre-analyzed semantic representations of candidate sentences from its retrieval backend. By means of methods for answer validation, it then tries to identify those candidate sentences that contain a correct answer to the question. There is a logic-based check for that, accepting an answer sentence as correct if the logical form of the parsed question can be proved from the MultiNet representation of the considered answer sentence (translated into a logical formula) and from the general background knowledge of LogAnswer. This check was made more robust by allowing query relaxation [FG10]. The addition of shallow linguistic features (such as lexical overlap) also serves to improve robustness. A reranking model based on rank-optimizing decision trees [Gl09] estimates the correctness probability of each answer candidate. Then, the system shows the five top-ranked answers to the user [GW12].

---

[2] http://www.loganswer.de/

## 3 Natural language representation based on MultiNet Graphs

The paradigm of Multilayered extended semantic networks (MultiNet) [He06] provides a semantic representation for natural language expressions. It also forms a comprehensive and thoroughly described knowledge representation system, based on a fixed repository of some 140 relations and functions whose properties are described by axioms. The concepts representations are enriched by multi-dimensional descriptions in terms of so-called layer attributes [He06]. These attributes serve to capture quantification and other aspects of meaning that cannot be expressed using the relational means of a semantic network. The layer attributes and their values will also be treated as part of the MultiNet graph. The formalism is generally independent of any particular natural language, although a full toolchain for translating NL into MultiNet is only available for German [He06]. The most important tool for my present purposes is the WOCADI parser [Ha03], which can translate a natural language expression into a MultiNet graph [He06]. Due to the use of MultiNet for semantic representation within LogAnswer, a CBR approach had to be developed which can handle such structured graph data. To choose MultiNet for the approach was a pragmatic decision. In this environment I already collected some experience with the paradigm and it fitted well with the idea to develop a CBR system with an optimized graph based similarity measure to improve the already existing tool set [He06],[GW12].

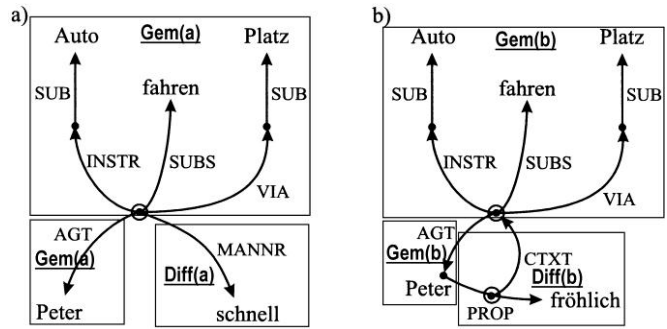

Figure 1: Example from [He06]) with division of similar and not similar graph components. The similar and not similar graph components are devised in sets Gem & Diff.

## 4 CBR Approach for MultiNet Graph classification

I enhance answer validation, by using experience knowledge in form of cases in a case base. This requires representing the experience knowledge in appropriate data structures. I further define a priori relations of current questions and answer candidates to already experienced answer candidates of former questions. The resulting system module is therefore for designed as a learning system and based on a Case Based Reasoning[3]

---

[3] CBR is an approach, for finding best matching problem solutions with similarity measures based on prior similar experiences. There are a lot of successfully running applications for problem solving in diagnosis,

control structure [Aa91], [We95], which addresses this problem solving task in particular. The main advantage of CBR systems is, that they can handle sparse, more noisy, or even absent information by retrieving the most similar case, if a complete match is not available in the case base. The Case Base is formally defined as in [Be02] with a case characterization part and a lesson part. Contrary to the usually expected approach in NLP I will not follow the textual approach, where experiences are available in unstructured or semi-structured text form [LA98]. In fact I use a structural approach with a common structured vocabulary [Be02]. The vocabulary container consists of XML describing MultiNet graphs [He06]. The explicit knowledge source of the CBR module [We95] is expert knowledge in the form of MultiNet graphs, bulk loaded and parsed from XML input files. These XML files where generated from the IRSAW module [GH07] and the CLEF 2007 and 2008 raw data.

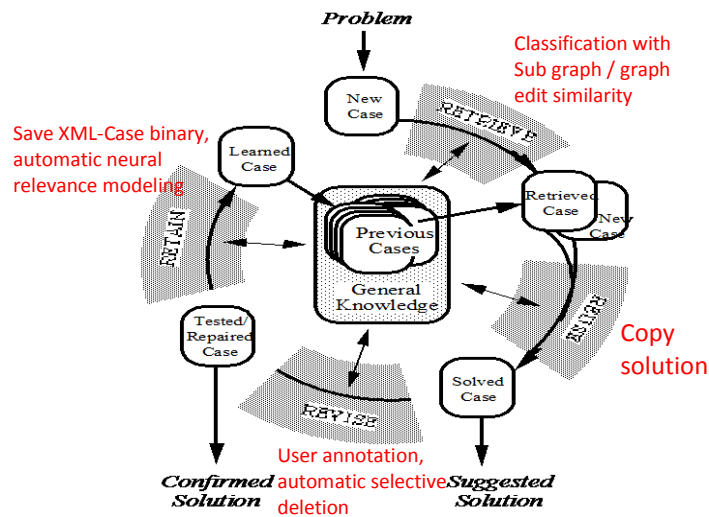

Figure 2: CBR Cycle by Aamodt and Plaza [Aa91], [AP94] and actions taken [We95],[GW12] by the presented approach.

**4.1 Definition of the Case Base**

The case base is formally defined as in [Be02] as an experience base with the help of MultiNet graphs [He06]. In my structural case representations, the describing features of a case are organized as MultiNet graph structures:

**Definition 1. (Space of Experience Characterization Descriptions)**: The *space of experience characterization[4] descriptions* D is the set of possible characterizations for experience items.

---

classification and decision support application domains [We95],[Al97],[Be02] e.g. for fault diagnosis, sales support, recipes, mission planning, design or project management.

[4] The characterization part describes the experience in a way that allows to assess its reusability in a particular situation [Be02].

$D = Q \times A$ *is a set of MultiNet graphs Q representing questions and A as a set of MultiNet graphs representing answer candidates.*

**Definition 2. (Space of Lessons):** The symbol L denotes the *space of possible lessons* recorded in experience items. In this approach *L is a Boolean denoting if the answer candidate contains a correct answer passage* [GP09].

**Definition 3. (Case, General Definition):** A *case* is a pair $c = (d, l) \in D \times L$.

**Definition 4. (Vocabulary Container and Case Space, General Definition):** We call the pair **VOC** = (D,L) the *vocabulary container*. The related *case space* is $C = D \times L$.

**Definition 5. (Case Base)** A *case base* is a finite set of cases, **CB** = *{c1, . . . , cn}* $\subseteq$ C.

## 4.2 Similarity measure

The core of the implicit knowledge modeling of the CBR module is the similarity measure. The retrieval engine works with a graph based similarity measure newly developed from standard graph similarity measures from [Be02] and adapted to the MultiNet methodology [He06]. I developed the new "integrated sub graph / edit similarity" for the purpose of answer validation in the NLP/MultiNet environment as follows.

### 4.2.1 Largest common sub graph matching similarity

One important component of the new integrated similarity measure is a graph matching similarity. Originally designed for not attributed graphs I distinguish among graph isomorphisms, sub graph isomorphisms, and the largest common sub graph measure. Where the isomorphic measures focus on bijective relations, the largest common sub graph approach allows us defining a non binary similarity measure.

**Definition 6. Largest Common Sub Graph [Be02]:** A graph *G* is the *largest common sub graph* of two graphs *G*1 and *G*2 (we write $G = lcsg(G1,G2)$), if $G \subseteq G1$ and $G \subseteq G2$ holds and there does not exist a graph *G'* such that $G' \subseteq G1$ and $G' \subseteq G2$ and $|G'| > |G|$. Here, $|(N,E)| = |N|+|E|$ is the size of the graph determined by the number of nodes and edges[5].

A symmetric[6] version of the largest common sub graph similarity measure can be defined as follows. It determines the similarity based on the relation between the size of the largest common sub graph and the maximum of the sizes of both graphs [Be02].[7]

$$sim_{lcsg}(x,y) = f(1 - \frac{|lcsg(x,y)|}{\max\{|x|,|y|\}})$$

---

[5] |(N,E)| = |N|+|E| = |G| is the size of the graph determined by the number of nodes and edges for the Largest Common Sub-Graph Measure. This is not valid for the following definition of the integrated sub graph / edit similarity measure.

[6] sim(x,y) = sim(y,x), the symmetric property is a not mandatory property, that nevertheless have a lot of similarity measure in common.

[7] f is a monotonic decreasing mapping function from [0, 1] to [0, 1] with f(0) = 1.

*4.2.2 Graph edit similarity*

First I determine the distance by counting and weighting the number of transformations necessary to transform one graph into the other [BM94].

$$\delta_{edit}(x,y) = \min\{\sum_{i=1}^{k} c(e_i) | (e_1, \ldots, e_k) \; transforms \; x \; to \; y\}$$

From this distance measure it is easy to derive the respective similarity measure:

$$sim_{edit}(x, y) = 1 - f(\delta_{edit}(x,y))$$

The cheaper and the fewer the operations are that are required to make the two graphs identical the smaller is the difference and hence the higher is the similarity [Be02].

*4.2.3 Integrated sub graph / edit similarity*

The similarity measure I use, profits by taking the advantages of both, the largest common sub graph matching similarity and the graph edit similarity. It mainly consists of the following: I measure the similarities of corresponding graph components e.g. attributes, lists of attributes, nodes and edges with local similarities [We95]. If I find pairs of components with a similarity higher than a threshold I put them into a set of sub graphs and graph components Gem, else in Diff. Then I sum up for each Graph component in Gem and Diff a certain relevance $\omega_k$ and compute the global similarity from it [We95]. Like the common sub graph- and the graph edit similarity measure [BM94], also the computation of the integrated sub graph / edit similarity measure is np-complete. Therefore the experimental system runs on an 8 Core 2.6GHz Xeon System in batch mode. Nevertheless test runs take several weeks.

**Definition 7**. Integrated subgraph graph edit similarity:

(i) With $x_i, y_i \subseteq$ Knots(X,Y) : analogue Edges

$$Sim_{op}(x_i,y_i) = sim_{Knots}(x_{io},y_{ip}) : o,p=1\ldots\#Knots \; in \; x,y$$

(ii) With $x_i, y_i \subseteq$ Lists(X,Y):

$$sim_{list}(x_i, y_i) = \frac{|x_i \cap y_i|}{\max(|x_i|, |y_i|)}$$

(iii) And for attributes with

$$sim_{attribute}(u,v) \begin{cases} 0 \leftrightarrow u \neq v \\ 1 \leftrightarrow u = v \end{cases}$$

Let then Gem, Diff $\subseteq$ X,Y be sub graphs or MultiNet graph features or components, $S_q$ relevance limits, and weights $\omega_k$ :

$$\forall q: sim_q(x_q, y_q) \geq S_q \to x_q \in Gem_{x_q}, y_q \in Gem_{y_q}$$

$$\forall q: sim_q(x_q, y_q) < S_q \rightarrow x_q \in Diff_{x_q}, y_q \in Diff_{y_q}$$

$$|Gem_{x_q}| = \sum_{k=0}^{\#Elements\ in\ Gem_{x_q}} \omega_k ,$$

analog $Gem_{y_q}, Diff_{x_q}, Diff_{y_q}$

Thus, the Integrated sub graph / edit similarity measure:

$$\text{Sim}_{\text{isg/g-es}}(X,Y) = \frac{\frac{|Gem_{x_q}|}{|Gem_{x_q}|+|Diff_{x_q}|} + \frac{|Gem_{y_q}|}{|Gem_{y_q}|+|Diff_{y_q}|}}{2}$$

The new integrated subgraph / graph edit similarity measure was designed to benefit from the obvious strengths and to abandon the obvious weeknesses of the largest subgraph and the rather complementary graph edit similarity measure.

## 5 Experimental results

The data set used for the experiments was generated by retrieving answer candidates for questions in the QA@CLEF 2007 and QA@CLEF 2008 test sets for German.[8] Due to the focus of CBR on the semantic network level, anaphoric references which are not of interest for the present purposes were resolved by filling in the antecedents and I only kept questions and answer candidates for which the network construction worked. I only kept questions with at least one correct answer candidate, otherwise a reranking is pointless. The annotation of all answer candidates for correctness with respect to answering the question was done manually. Finally, duplicate answer candidates were removed. This left us with 254 questions and more than 15,000 items in the annotated data set.[9,10] The systems capabilities so what rise and fall with the quality of the parses from natural language into MultiNet and the quality of user annotations.

---

[8] See http://www.clef-initiative.eu/ for more information about the CLEF system evaluation campaign.
[9] Note that due to these restrictions, we could not extend the experiment to the full QA@CLEF 2007 and 2008 test sets. The LogAnswer system has fallback techniques to handle the 146 questions that were taken aside, but they are not of interest here since our back-off approach does not involve CBR, yet.
[10] Unfortunately a full CLEF Setup was not possible due to the following reasons:
1. The IRSAW Component was not able to transform all given sentences into MultiNET graphs.
2. It did not make sense to leave cases with questions with only false and no correct answer candidates in the case base.
3. There was no relevant reason to leave duplicates in the case base. In fact a case base should be as small as possible to enhance the performance of the runtime system [We95]. Further duplicates lead to a correct complete match and falsify the results.

**5.1 Case Retrieval Experiments**

I measured the system classification accuracy by running tests with a case base constructed from the questions and answer candidates described in the previous section. For the optimization task I used simple hill climbing algorithms. I rejected the optimal A* approach only because of its calculation complexity. I conducted several experiments, namely:

1. Retrieve the same meaning or synonym question / answer candidate pairs. As there is for every query a very similar correct best match in the case base 100% are correctly retrieved.
2. Retrieve already asked questions with new answer candidate pairs. The system finds the question and retrieves in 73% a correct classification (56% for correct cases). In this test setting the integrated subgraph/graph edit similarity measure outperformed the largest subgraph similarity measure 34% (3% for correct cases) and the graph edit similarity measure 27% (2% for correct cases). So with these no more experiment were made. The results far below the at least expected ca. 50% are probably due to the heavily unbalanced test data.
3. Retrieve new questions with new answer candidate pairs. Here I temporarily deleted the probed query question from the case base. The system therefore cannot find a very similar best match. The result fell to 59% (29% for correct answers).
4. Retrieve new questions with new answer candidate pairs with a case base optimized to get optimal results for the classification of correct questions / answer candidate pairs, of already known questions / answer candidate pairs. I deleted with a simple hill climbing[11] algorithm incorrect cases disturbing the classification of the correct cases (questions with correct answer candidates). The classification of correct cases rose then to 100% (overall: 82%).
5. Retrieve new questions with new answer candidate pairs with a case base optimized to get optimal overall results, of already known questions / answer candidate pairs. I now deleted with a simple hill climbing correct cases, disturbing the overall result. The result now rose to 87% (96% for correct answers). This is the optimal result achieved with the current setup for questions, not already experienced within the case base. One can see, that the result degrades with the less experience accumulated in the case base.
6. Retrieve new questions with new answer candidate pairs with a raw case base, in a 3-fold cross reference test. This was the same setup as 3 with a one third smaller case base as a base line for comparison with setup 7 and 8. I got a 56% classification rate on the test folds (23% for correct answers).
7. Retrieve new questions with new answer candidate pairs with a case base optimized to get optimal results for the classification of correct questions, in a cross reference test. One can see as expected a smaller effect of the hill climbing optimization than in 4. The classification rate is 56% overall, 40% for correct answers.
8. Retrieve new questions with new answer candidate pairs with a case base optimized to get optimal overall results, in a 3-fold cross reference test. As in 7 one can see again a smaller effect of the optimization. The classification rate is 55%, 29% for correct answers.

---

[11] The hill climbing deletes only cases not complete questions. The number of questions covered by the case base remained the same.

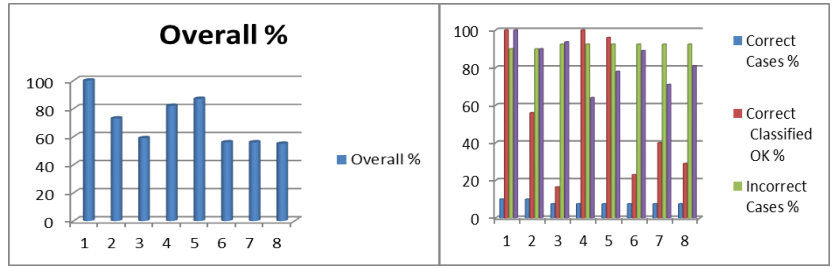

Figure 3,4: Overall percentage (mean value of correct and incorrect case queries results) results of the 8 test rows and detailed percentage results' comparison.

**5.2 User interaction simulation**

I performed a test to examine the results' development with an increasing number of correctly annotated Cases in the Case-Base. I started with a case base of 1000 cases and augmented the number of cases for each step by about 1000 cases by complete questions. For each step I completed the simple hill climbing optimization[12].

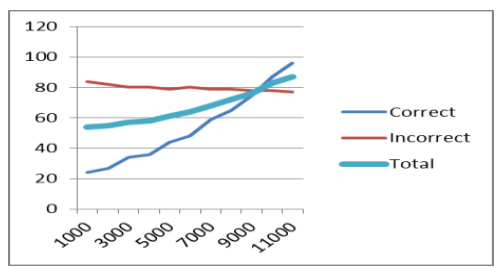

Figure 5: Percentage comparison of the user interaction simulation.

The results show the increase of the overall retrieval's accuracy with a growing number of correct cases in the case base.

**5.3 The machine learning integration experiment**

In the QA setting, answer reranking is the art of shifting correct answer candidates for a question to top ranks, by assessing their likelihood of being correct. This paper aims at improving a solution for answer reranking by incorporating CBR [GW12], so that the running QA system can profit from every new user annotation extending the case base. In order to be able to integrate case-based reasoning and existing answer selection techniques, the results of the CBR stage are turned into numeric features. A ranking model determined by a supervised learning-to-rank approach combines these CBR-based features with other answer selection features determined by shallow linguistic processing and logical answer validation [FG10]. Rank-optimizing decision trees [Gl09] are used for learning an answer reranking model from the existing answer validation features of LogAnswer [FG10] and the new CBR features. This supervised learning method was specifically developed for imbalanced data, such as typical sets of answer candidates that contain only a small fraction of correct answers. For the existing feature set, We adopt the monotonicity specifications documented in [FG10]. The final ML ranker is an ensemble of ten rank-optimizing decision trees, obtained by stratified bagging, whose individual probability estimates are combined by averaging. Stratified bagging serves to learn several trees from a given training set and to combine them into an ensemble classifier [Gl09]. Each tree is pruned to a size of 40 splits, which turned out to be suficient for the task at hand. As was shown in [Gl09], bagging of rank-optimizing decision trees outperforms usual decision-tree learners (and other common ML techniques) in answer selection tasks.

---

[12] The numbers are rounded, because we always added cases by question. So we started the optimization process with more than a plain thousand number. During following optimization process the hill climbing algorithm deleted more or less the same number of cases.

*5.3.1 Answer Reranking Experiments Based on CBR Features - Evaluation of the Learning-to-Rank Approach*

The main results of the learning-to-rank approach for various combinations of feature sets are shown in Figure 6 (ML results for case baseoptimized for treatment of correct candidates). The feature sets used for training the ML models were D: deep features (based on results of logic-based answer validation, as described in [FG10]), I: the original retrieval score of the candidate retrieval system, S: shallow features (based on a lexical overlap check and similar shallow criteria, also described in [FG10]), and C: CBR features. The number 3 indicates the limit on allowable relaxation steps in the proof attempts.

| Model | MRR | ANS-1 | ANS-2 | ANS-3 | ANS-4 | ANS-5 |
|---|---|---|---|---|---|---|
| DSC3 | **0.74** | **0.61** | **0.76** | **0.83** | **0.88** | **0.89** |
| DS3 | 0.72 | 0.58 | **0.76** | **0.83** | 0.87 | **0.89** |
| S | 0.71 | 0.58 | 0.74 | 0.82 | 0.86 | 0.90 |
| SC | 0.69 | 0.55 | 0.72 | 0.80 | 0.85 | 0.86 |
| irScore | 0.48 | 0.33 | 0.46 | 0.55 | 0.62 | 0.67 |
| CI | 0.37 | 0.22 | 0.34 | 0.41 | 0.47 | 0.51 |
| C | 0.27 | 0.15 | 0.24 | 0.30 | 0.34 | 0.37 |

Figure 6: ML Results for Case Base Optimized for Retrieval of Correct Candidates, sorted by MRR

As shown by Figure 6 when training the ML ranker on a case base optimized for perfect treatment of correct answer candidates, I get the best overall result in our tests, with an MRR of 0.74 and a correct top-ranked answer chosen in 61% of the cases. It is instructive to consider the usage of CBR features in the best ML ranker DSC3 shown in Figure 7, by inspecting all branching conditions in the generated trees and counting the frequency of occurrence of each feature in such a branching condition (since 10 bags of 10 decision trees each were generated in the 10 cross-validation runs, there is a total of 100 trees to base results on) [GW12]. In total, 42.5% of all split conditions in the learned trees involve one of the CBR attributes. This demonstrates a strong impact of CBR results on answer reranking, so that new experience in the case base will also become effective in answer reranking [GW12].

# 6 Conclusions and future work

I have presented an integrated CBR approach to answer validation and reranking for QA systems. Since each annotated answer candidate for a user question provides evidence for validating answer candidates for future questions, this approach promises a continuous increase in answer quality, given a feedback mechanism for users that extends the case base. In the paper, I have emphasized the use of CBR techniques, namely the structural case base, built with annotated MultiNet graphs, and corresponding graph similarity measures. I have covered a priori relations to experienced answer candidates for former questions, and I have structured the case base and defined the relevant similarity measures in a way suitable for solving the answer validation problem. Finally I have integrated CBR into the existing learning-to-rank approach by defining CBR-related features. Data from QA@CLEF tasks served to evaluate case retrieval and the achieved reranking quality. Judging from classification rate, it works extremely well for question/answer candidate pairs of the same meaning, and still quite good if the question was already asked and correctly answered, even if the correct answer candidate is of a different semantic construction. The less similar the questions and answer candidates become to already experienced ones, the worse the quality of the best matches. This is also the range within which the already tested optimizations work fine, as shown by the CBR tests based on cross validation. Concerning the integration of CBR results into the existing answer reranking solution, experiments confirm that the best learned models include CBR features, achieving an MRR up to 0.74 with a correct top-ranked answer shown in 61% of the cases. There is already an advantage over the original approach without CBR, which will increase with time due to the automatic improvements with new user annotations. The massive use of CBR features by the learning- to-rank method, comprising 42.5% of all branching conditions in the learned decision trees, shows that the CBR features have a strong impact on answer reranking - and this is the prerequisite for improvements in the case base to translate into corresponding improvements of ranking results. The working overall architecture offers a point of departure for future improvements. Thus, I will explore improvements of the similarity measure on MultiNet graphs. I will determine the best classifying local similarity measures for the task. For that purpose, I need to analyze in depth the local similarity results to enhance the classification results. Further I will optimize the structure and weights of the global similarity measures to optimize the results in areas where I currently retrieve only 50-60% correct best matches overall, namely the question/answer candidate pairs completely different from the case base. Another option will be to analyze in depth the most similar components of the MultiNet graphs of correct question/answer pairs to find generalizations or abstractions that will further help to classify those question/answer candidate pairs that do not share the specific lexical concepts of cases in the case base, but rather show parallels in the MultiNet graph structure. To this end, I shall integrate more explicit knowledge from the reasoning process and create effective similarity measures on those to enhance the retrieval sub process. I aim to optimize the similarity measure by using more adapted, relevant implicit knowledge and integrate more explicit knowledge [We95]. I will try to profit from generalized or abstracted cases [Be02] to get a more general view within the elaboration of the similarities.